\documentclass{article}

\usepackage{microtype}
\usepackage{graphicx}
\usepackage{subfigure}
\usepackage{booktabs}
\usepackage{hyperref}

\usepackage[arxiv]{icml2023}
\usepackage{amsmath}
\usepackage{amssymb}
\usepackage{mathtools}
\usepackage{amsthm}
\usepackage[capitalize,noabbrev]{cleveref}

\theoremstyle{plain}

\theoremstyle{definition}

\theoremstyle{remark}

\usepackage[textsize=tiny]{todonotes}

\usepackage{diagbox}
\usepackage{makecell}
\usepackage[normalem]{ulem}

\newcommand{\tablestyle}[2]{\setlength{\tabcolsep}{#1}\renewcommand{\arraystretch}{#2}\centering\small}
\newcommand{\tuner}{{\color[RGB]{0,0,0}\texttt{\textbf{U-Tuner}}}}
\newcommand{\utuning}{{\color[RGB]{0,0,0}\texttt{U-Tuning}}}
\newcommand{\uop}{{\color[RGB]{0,0,0}\text{OP}}}

\definecolor{myMagenta}{rgb}{0.9,0,0.4}

\icmltitlerunning{Rethinking Efficient Tuning Methods from a Unified Perspective}
\begin{document}

\twocolumn[
\icmltitle{Rethinking Efficient Tuning Methods from a Unified Perspective}

\begin{icmlauthorlist}
\icmlauthor{Zeyinzi Jiang}{ali}
\icmlauthor{Chaojie Mao}{ali}
\icmlauthor{Ziyuan Huang}{nus,ali}
\icmlauthor{Yiliang Lv}{ali}
\icmlauthor{Deli Zhao}{ali}
\icmlauthor{Jingren Zhou}{ali}
\end{icmlauthorlist}

\icmlaffiliation{ali}{Alibaba Group}
\icmlaffiliation{nus}{National University of Singapore}

\icmlcorrespondingauthor{Zeyinzi Jiang, Chaojie Mao, Yiliang Lv}{zeyinzi.jzyz, chaojie.mcj, yiliang.lyl@alibaba-inc.com}
\icmlcorrespondingauthor{Ziyuan Huang}{ziyuan.huang@u.nus.edu}
\icmlcorrespondingauthor{Deli Zhao}{zhaodeli@gmail.com}
\icmlcorrespondingauthor{Jingren Zhou}{jingren.zhou@alibaba-inc.com}

\icmlkeywords{Parameter-efficient Transfer Learning, Vision Transformer, Unified Architecture}

\vskip 0.3in
]
\printAffiliationsArxiv{}

\begin{abstract}

Parameter-efficient transfer learning (PETL) based on large-scale pre-trained foundation models has achieved great success in various downstream applications. Existing tuning methods, such as prompt, prefix, and adapter, perform task-specific lightweight adjustments to different parts of the original architecture. However, they take effect on only some parts of the pre-trained models, \textit{i.e.,} only the feed-forward layers or the self-attention layers, which leaves the remaining frozen structures unable to adapt to the data distributions of downstream tasks. Further, the existing structures are strongly coupled with the Transformers, hindering parameter-efficient deployment as well as the design flexibility for new approaches. In this paper, we revisit the design paradigm of PETL and derive a unified framework \utuning\ for parameter-efficient transfer learning, which is composed of an operation with frozen parameters and a unified tuner that adapts the operation for downstream applications. The \utuning\ framework can simultaneously encompass existing methods and derive new approaches for parameter-efficient transfer learning, which prove to achieve on-par or better performances on CIFAR-100 and FGVC datasets when compared with existing PETL methods.

\end{abstract}

\section{Introduction} \label{sec:intro}

\begin{figure}[ht]
\vskip 0.1in
\begin{center}
\centerline{\includegraphics[width=\columnwidth]{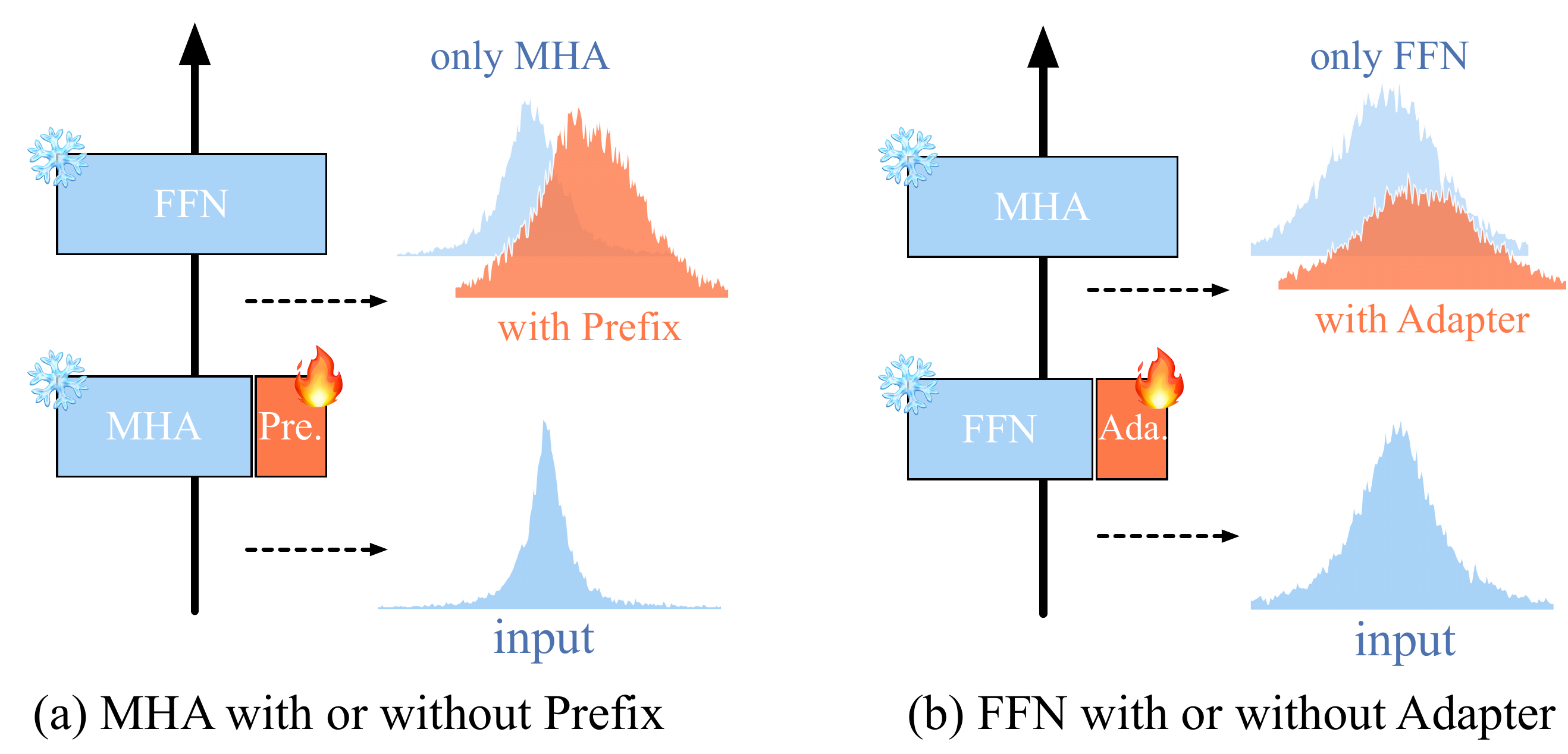}}
\caption{
Visualization of input and output data distributions (a) in the first Transformer block and (b) between the first and the second Transformer block. Existing PETL methods only adjust the distribution of some parts of the pre-trained Transformer models, which brings difficulty for the remaining frozen model parts to adapt to the new distribution.
}
\label{fig:distribution}
\end{center}
\vskip -0.4in
\end{figure}

The increasingly large amount of data and powerful computing resources have driven the emergence of foundation models with large  capacities~\cite{devlin2018bert,brown2020gpt3,radford2021clip,jia2021align}, which demonstrate strong generalization ability across multiple downstream tasks in visual~\cite{he2022mae,bao2021beit}, language~\cite{liu2019roberta, 2020t5} and multi-modal paradigms~\cite{li2019visualbert,yu2022coca,alayrac2022flamingo}. On the flip side, the increasing size of pre-trained models has also made the training, storage, and deployment cost incredibly high for downstream adaptation.

To address this issue, the recent efforts are made in parameter-efficient transfer learning (PETL)~\cite{houlsby2019adapter}, which proves to perform competitively against the fully-fine-tuning counterparts~\cite{karimi2021compacter,jia2022vpt}. Existing approaches for PETL either tune a few existing parameters~\cite{zaken2021bitfit} or insert additional trainable structures~\cite{lester2021pompt, li2021prefix, hu2021lora} with most or all pre-trained parameters frozen, maintaining a low training and deployment cost for downstream adaptation. 

Despite the strong performance, the current practice has two shortcomings. \textbf{(i)} Most existing approaches insert learnable structures to only some parts of the model, leaving the remaining frozen parts struggling to adapt to the new distribution generated by the tuned blocks. For example, introducing prefix tokens~\cite{li2021prefix} essentially changes the output distribution of multi-head attention (MHA) in pre-trained Transformers, while the subsequent feed-forward network (FFN) is trained for the original distribution and will thus encounter difficulty in adapting to the new distribution. This goes for inserting trainable structures in FFNs as well, as can be observed in~\cref{fig:distribution}. \textbf{(ii)} The structure of existing tuning methods and the main branch of the pre-trained models are strongly coupled, which means most existing structures are deeply embedded in the Transformers. On one hand, this hinders parameter-efficient deployment since the pre-trained model and the tuning structures need to be deployed entirely for every downstream application. On the other hand, the strong degree of coupling also limits the design flexibility for new approaches for PETL.

\begin{figure}[t]
\vskip 0.1in
\begin{center}
\centerline{\includegraphics[width=0.8\columnwidth]{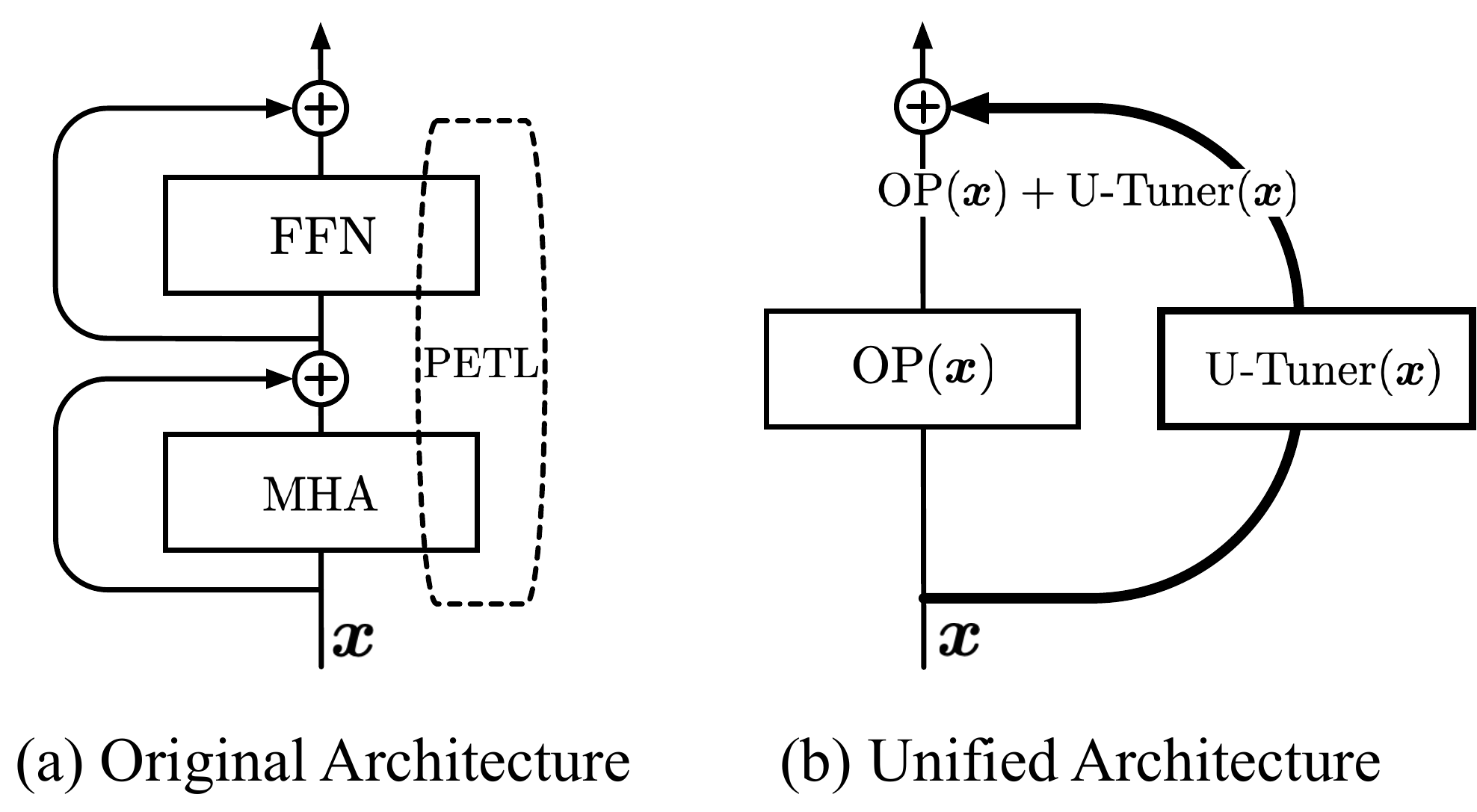}}
\caption{
The diagram of existing methods and our unified framework for PETL. (a) Existing PETL methods are deeply embedded into original structures. (b) Our \utuning\ framework is composed of a frozen operation \uop\ and a unified tuner \tuner.
}
\label{fig:sketch}
\end{center}
\vskip -0.5in
\end{figure}

In this work, we rethink parameter-efficient transfer learning from a unified perspective towards existing approaches. Specifically, we revisit existing tuning paradigms and found a parallel form for mainstream tuning methods, such as adapters~\cite{houlsby2019adapter}, prefix tuning~\cite{li2021prefix}, and prompt tuning~\cite{lester2021pompt}, which reduces the degree of coupling for tuning structures. This ability to transform to a parallel form for most mainstream methods also allows us to reveal a unified formulation. Based on this, we provide a unified framework for parameter-efficient transfer learning, which we call \utuning. Composed of an operation with frozen parameters and a lightweight unified trainable structure (as in \cref{fig:sketch}), the \utuning\ framework allows for flexible insertion of tuning structures. Hence, it can not only encompass most existing works but also allows for the derivation of new tuning structures. Extensive experiments on transfer learning show the generality of the \utuning\ framework, and the derived new PETL models prove to achieve on-par or better performances on various downstream tasks. 

Our contributions can be summarized as follows: \textbf{(i)} We derive a parallel form for mainstream PETL methods, which reduces the degree of coupling and facilitates parameter-efficient deployment of large pre-trained models. \textbf{(ii)} We propose a unified tuning framework \utuning\ that encompasses existing PETL methods and allows for the derivation of new ones. \textbf{(iii)} Comprehensive studies on transfer learning prove the generality and powerfulness of \utuning.

To the best of our knowledge, this is the first work unifying PETL methods in a backbone-independent form, providing a different perspective for efficient tuning.

\section{Related Work} \label{sec:related}

\textbf{Transformers in computer vision.}
Transformers~\cite{vaswani2017attention} have achieved great success in various fields~\cite{brown2020gpt3,ramesh2022dalle2}. In computer vision, Vision Transformers (ViT)~\cite{dosovitskiy2020vit} are widely applied in various vision tasks, \textit{e.g.}, visual classification~\cite{liu2021Swin,li2022uniformer}, object detection~\cite{song2022vidt,carion2020detr} and segmentation~\cite{SETR,wang2021PVT}, and have demonstrated strong generalization ability when pre-trained on a large corpus of data. A typical ViT consists of a cascade of Transformer blocks in which each block is constructed with a multi-head attention (MHA) and a feed-forward network (FFN). The existing PETL methods mainly perform lightweight adjustments to particular parts in the MHA, FFN, or the whole block.

\textbf{MHA-based tuning.}
MHA-based tuning embeds trainable parameters in MHA. LoRA~\cite{hu2021lora} constructs an additional layer with low-rank decomposition matrices of the weights in the network. Derived from specific textual templates~\cite{brown2020gpt3, liu2021pre}, prompt tuning~\cite{lester2021pompt} prepends extra trainable tokens~\cite{lester2021pompt, liu2021gpt, Liu2021ptuningv2} to the input. Prefix tuning~\cite{li2021prefix} optimizes the task-specific vector in the multi-head attention layer. In computer vision, visual prompt tuning (VPT)~\cite{jia2022vpt} is proposed to initialize tunable prompt tokens and prepend to the original tokens in the first layer or multiple layers.

\textbf{FFN-based tuning.}
For FFNs, the adaptation is generally made by adapter~\cite{houlsby2019adapter} and its generalized versions~\cite{pfeiffer2020adapterfusion, karimi2021parameterefficient,karimi2021compacter,he2022towards}, which usually insert a bottleneck layer into each FFN layer. In the video paradigm, adapters can also be used to introduce temporal information in FFNs~\cite{chen2022adaptformer}.

\textbf{Block-based tuning.}
Block-based tuning updates the common term of the block. A simple solution is BitFit~\cite{zaken2021bitfit}, which only tunes the bias terms of the model. Diff pruning~\cite{guo2020diff} learns a diff vector, which is adaptively pruned during training process.

\textbf{Other tuning.}
In addition, some of the efforts are made in finding out the optimal design paradigm of tuning modules. NOAH~\cite{zhang2022NOAH} attempts to find out the optimal design of tuning modules through a neural architecture search algorithm. \citeauthor{he2022towards} design a variety of forms about previous methods in a unified view and validate the best choices of variant adapters empirically. 

In this paper, we rethink and analyze the previous methods systematically and propose a unified tuning framework, which presents to be a flexible and extensible paradigm for PETL on various downstream tasks.

\begin{figure*}[t]
\vskip 0.1in
\begin{center}
\centerline{\includegraphics[width=2\columnwidth]{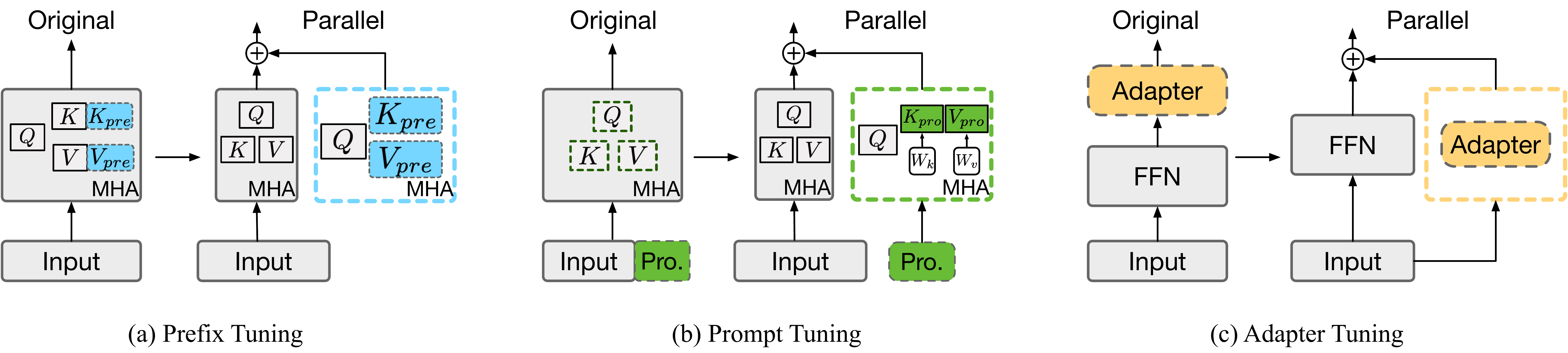}}
\caption{
Illustration of (a) prefix tuning~\cite{li2021prefix}, (b) prompt tuning~\cite{lester2021pompt}, and (c) adapter tuning~\cite{houlsby2019adapter} for parameter-efficient transfer learning and their parallel form. 
}
\label{fig:parallel}
\end{center}
\vskip -0.2in
\end{figure*}

\section{Rethinking PETL} \label{sec:rethink}

In this section, we take a closer look into several existing approaches for PETL. Interestingly, we found that all of them can be equivalently formulated as the parallel combination of (i) existing operations (\textit{e.g.}, multi-head attention (MHA) or feed-forward networks (FFN)) in Transformer architecture, and (ii) newly introduced structures performing a similar operation. Our investigation in these approaches is divided into two groups, MHA-based tuning methods, and FFN-based tuning methods.

\subsection{MHA-based Tuning}
Multi-head attention (MHA) based tuning usually takes effect on the self-attention part of the Transformer blocks. The original self-attention~\cite{vaswani2017attention} is expressed as:
\begin{equation}
\small
\label{eq:attn}
    \operatorname{Attn}(\boldsymbol{Q}, \boldsymbol{K}, \boldsymbol{V})=\operatorname{softmax}\left(\frac{ \boldsymbol{Q} \boldsymbol{K}^{\mathrm{T}}}{\sqrt{d}}\right)V,
\end{equation}
where $\boldsymbol{Q}$, $\boldsymbol{K}$ and $\boldsymbol{V}$ denote the query, key and value respectively. Usually, given input tokens $\boldsymbol{x}$, the query, key, and value are obtained through a linear projection $\boldsymbol{Q}=\boldsymbol{x}\boldsymbol{W}_{q}$, $\boldsymbol{K}=\boldsymbol{x}\boldsymbol{W}_{k}$, and $\boldsymbol{V}=\boldsymbol{x}\boldsymbol{W}_{v}$, where $\boldsymbol{W}_q$, $\boldsymbol{W}_k$ and $\boldsymbol{W}_v$ are learnable projection weights.

\textbf{Prefix tuning.}
To adapt the output distribution of MHA to downstream tasks, prefix tuning~\cite{li2021prefix} keeps the projection weights in self-attention unchanged and prepends learnable parameters $\boldsymbol{K}_{pre}$ and $\boldsymbol{V}_{pre}$, \textit{i.e.,} prefix tokens, to the projected key and value respectively:
\begin{equation}
\label{eq:prefix}
\begin{array}{l}
\operatorname{MHA}_{\text{pre}} = \operatorname{Attn}(\boldsymbol{x}\boldsymbol{W}_{q}, [\boldsymbol{K}_{pre}; \boldsymbol{x}\boldsymbol{W}_{k}], [\boldsymbol{V}_{pre}; \boldsymbol{x}\boldsymbol{W}_{v}]).
\end{array}
\end{equation}
Essentially, when we view the $\operatorname{MHA}_{\text{pre}}$ as performing MHA separately between the query and the original keys and values, and between the query and the prefix tokens, we can obtain an equivalent form as:
\begin{equation}
\small
\label{eq:prefix_parallel}
\begin{array}{l}
\operatorname{MHA}_{\text{pre}} = (1-\lambda) \underbrace{\operatorname{Attn}\left(\boldsymbol{Q}, \boldsymbol{K}, \boldsymbol{V}\right)}_{\text {original attention }}+ 
\lambda\underbrace{\operatorname{Attn}\left(\boldsymbol{Q}, \boldsymbol{K}_{pre}, \boldsymbol{V}_{pre}\right)}_{\text{prefix attention in parallel}},
\end{array}
\end{equation}
where $\lambda$ weighs between the original attention and prefix attention in parallel. Detailed value for $\lambda$ and the derivation process are included in~\cref{app:form}.

In this way, the original MHA, $\operatorname{Attn}(\boldsymbol{Q}, \boldsymbol{K}, \boldsymbol{V})$ and the MHA for prefixes, $\operatorname{Attn}(\boldsymbol{Q}, \boldsymbol{K}_{pre}, \boldsymbol{V}_{pre})$ can be computed in parallel. Hence, we call \cref{eq:prefix_parallel} the parallel form of prefix tuning. Both the original form and the parallel form of prefix tuning can be seen in~\cref{fig:parallel} (a).

\textbf{Prompt tuning.}
Instead of introducing learnable parameters at the same level as projected keys $\boldsymbol{K}$ and values $\boldsymbol{V}$, prompt tuning~\cite{lester2021pompt} introduces learnable latent tokens $\boldsymbol{x}_{pro}$, \textit{i.e.,} prompts, at the same level as the input tokens:
\begin{equation}
\small
\label{eq:prompt}
\begin{array}{l}
\operatorname{MHA}_{\text{pro}}  = \operatorname{Attn}\left([\boldsymbol{x}; \boldsymbol{x}_{pro} ] \boldsymbol{W}_{q}, [\boldsymbol{x}; \boldsymbol{x}_{pro} ] \boldsymbol{W}_{k}, [\boldsymbol{x}; \boldsymbol{x}_{pro} ] \boldsymbol{W}_{v} \right), 
\end{array}
\end{equation}
where the prompts $\boldsymbol{x}_{pro}$ are concatenated to the input token $\boldsymbol{x}$ in the first layer or multilayer.

Similar to prefix tuning, we can obtain the parallel form of prompt tuning as:
\begin{equation}
\small
\label{eq:prompt_parallel}
\begin{array}{l}
\operatorname{MHA}_{\text{pro}}
=[ (1-\lambda) \underbrace{ \operatorname{Attn}\left(\boldsymbol{Q}, \boldsymbol{K}, \boldsymbol{V}\right) }_{\text{original attention}}+ 
\lambda \underbrace{\operatorname{Attn}\left(\boldsymbol{Q}, \boldsymbol{K}_{pro}, \boldsymbol{V}_{pro}\right)}_{\text{prompt attention in parallel}};
\vspace{0.2cm}\\ 
(1-\beta)\operatorname{Attn}\left(\boldsymbol{Q}_{pro}, \boldsymbol{K}_{pro}, \boldsymbol{V}_{pro}\right)+ 
\beta\operatorname{Attn}\left(\boldsymbol{Q}_{pro}, \boldsymbol{K}, \boldsymbol{V} \right)
],
\end{array}
\end{equation}
where $\boldsymbol{K}_{pro}$ and $\boldsymbol{V}_{pro}$ are the key and value produced by prompt tokens, respectively. $\lambda$ and $\beta$ are individual attention weights. More details can be seen in~\cref{app:form}.

Note that~\cref{eq:prompt_parallel} concatenates the prompt tokens with the original input tokens. In practice, prompt tokens are often discarded after the MHA. In this case, the concatenation operation in~\cref{eq:prompt_parallel} can be ignored, and thus the $\operatorname{MHA}_{\text{pro}}$ can have a similar form as~\cref{eq:prefix_parallel}. The original form and the parallel form of prompt tuning can be seen in~\cref{fig:parallel} (b).

\subsection{FFN-based Tuning}
In Transformers~\cite{vaswani2017attention}, feed-forward networks (FFN) is a multi-layer perceptron (MLP) block directly connected to MHA, which is expressed as:
\begin{equation}
\label{eq:ffn}
    \operatorname{FFN}(\boldsymbol{x})={\phi}(\boldsymbol{x}\boldsymbol{W}_1+\boldsymbol{b}_1)\boldsymbol{W}_2+\boldsymbol{b}_2\ ,
\end{equation}
where $\boldsymbol{W}_1$ and $\boldsymbol{W}_2$ are the projection weights, $\boldsymbol{b}_1$ and $\boldsymbol{b}_2$ are the bias terms, and $\phi$ is the non-linear activation function between consecutive fully-connected layers.

\textbf{Adapter tuning.}
Adapter~\cite{houlsby2019adapter} is typically used to adjust the output distribution for FFN in Transformers. Usually, adapters project input tokens $\boldsymbol{x}$ by an MLP layer, similar to FFN. Differently, instead of expanding the feature dimension of the input tokens, adapters usually first shrink the channel dimension before it is expanded back to the original number:
\begin{equation}
\label{eq:adapter_parallel}
\operatorname{FFN}_{\text{adapter}} = \underbrace{ \operatorname{FFN(\boldsymbol{x})} }_{\text {original module}} + 
\underbrace{ {\phi}( \operatorname{FFN(\boldsymbol{x})} \boldsymbol{W}_{down}) \boldsymbol{W}_{up} }_{\text {adapter module in parallel} },
\end{equation}
where $\boldsymbol{W}_{down}$ and $\boldsymbol{W}_{up}$ denote the weights for the down-projection layer and the up-projection layer, respectively.

Essentially, the FFN module with adapters can be thought of as the parallel connection between the original FFN module and the FFN module with an MLP layer, \textit{i.e.,} an adapter\footnote{For clarity, the residual connection in the outer layer of FFN is not taken into account.}. The comparison between the structures of the original form and the parallel form can be seen in~\cref{fig:parallel} (c). Another way to insert adapters is to add a scaling factor and design the adapter explicitly as a parallel module~\cite{he2022towards,chen2022adaptformer}, which can be similarly viewed as parallel structures.

\begin{figure*}[htbp]
\vskip 0.1in
\begin{center}
\centerline{\includegraphics[width=1.7\columnwidth]{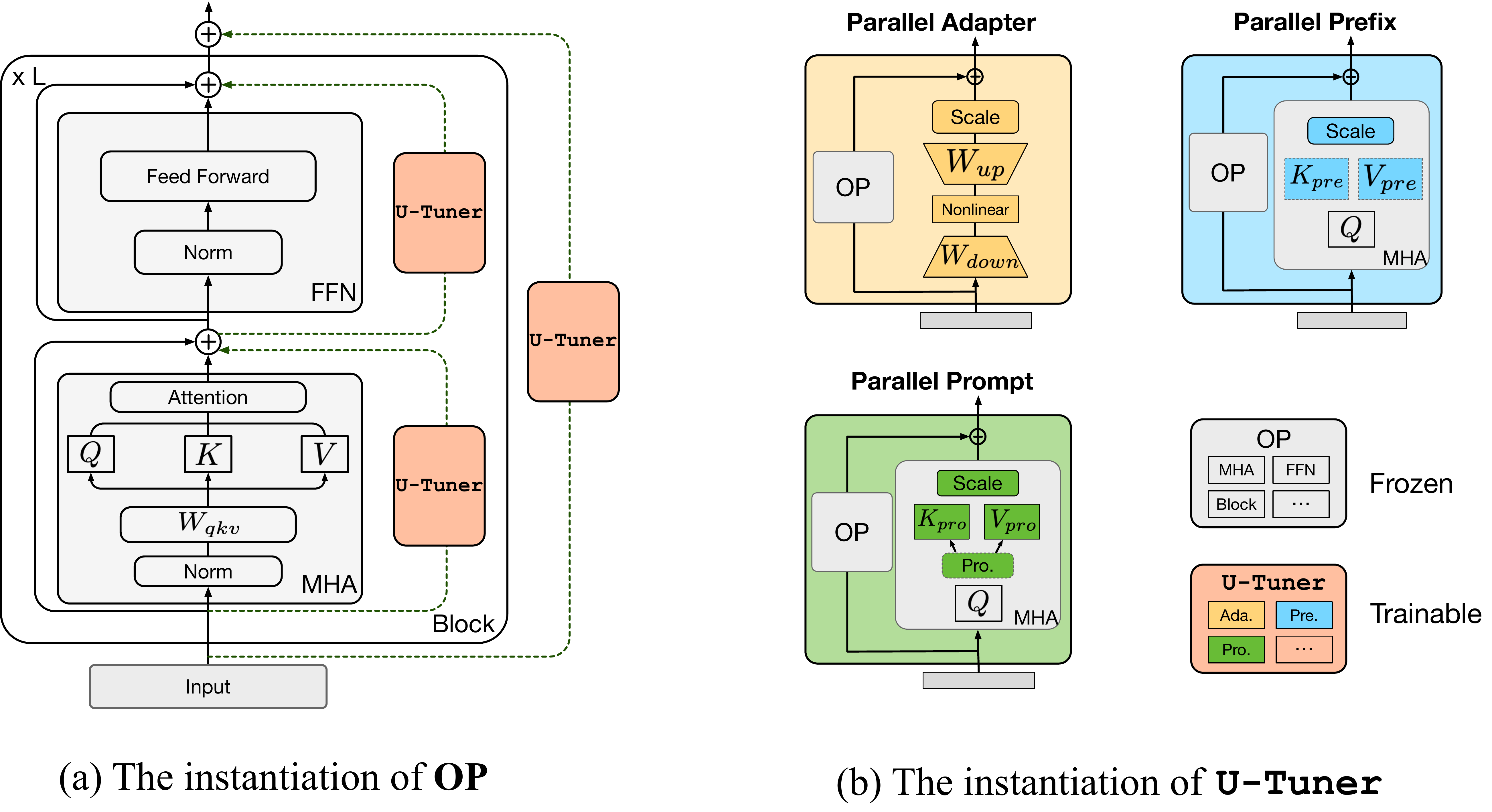}}
\caption{
Our \utuning\ framework for parameter-efficient transfer learning (PETL) consists of an operation (\uop) with frozen parameters and a unified tuner (\tuner) that adapts the output distribution to downstream applications. Unlike existing tuning methods, the \uop\ and the \tuner\ are instantiated in a decoupled way, which means the instantiation of the \uop\ and \tuner\ can have arbitrary combinations. (a) The instantiation of \uop\ can be MHA, FFN, and the whole block in pre-trained Transformers. (b) The instantiation of \tuner\ can be a parallel adapter, parallel prefix, and parallel prompt. Note that when new structures or operations emerge, our \utuning\ framework can always be flexibly extended to encompass new operations.
}
\label{fig:utuning}
\end{center}
\vskip -0.3in
\end{figure*}

\section{U-Tuning} \label{sec:utuning}
In this section, we first present a unified perspective on the existing tuning paradigms. Based on this, a \underline{U}nified \underline{Tuning} framework, dubbed  \utuning, is proposed for PETL, which is composed of a frozen operation existing in pre-trained Transformers and a learnable \tuner\  for adjusting the output distributions. After that, we instantiate our \utuning\ with various building blocks.

\subsection{A Unified Formulation} \label{sec:formulation}
From~\cref{eq:prefix_parallel,eq:prompt_parallel,eq:adapter_parallel}, we can make two observations: \textbf{(i)} existing tuning methods use a similar structure, \textit{i.e.,} the parallel combination of an existing operation with frozen parameters and a newly-introduced structure with learnable parameters. The frozen parts yield generalized representations learned from large-scale data, while the learnable parts adapt the generalized representations to specific downstream tasks. \textbf{(ii)} the learnable parts of existing tuning methods generally have a similar structure to the corresponding frozen parts, \textit{e.g.,} prefix for MHA essentially introduces self-attention between existing queries and prefixes.

Given the above observations, we generalize existing tuning methods for PETL into a unified formulation:
\begin{equation}
\label{eq:uni}
\boldsymbol{x}^{\prime} = \uop(\boldsymbol{x})  + \tuner(\boldsymbol{x}) 
\end{equation}
where $\boldsymbol{x}/\boldsymbol{x}^{\prime}$ are input/output tokens, \uop\ denotes the existing operations in Transformers with frozen parameters, and \tuner\ represents the unified tuner, which is newly-introduced for adjusting the output distributions of \uop.

Our unified formulation views each part of Transformer as an operation function \uop, while each tuned part as a unified tuner \tuner. In this way, the formulation encompasses all existing tuning methods when we instantiate \uop\ and \tuner\ with similar operations. Meanwhile, it enables the generation of new tuning methods for parameter-efficient transfer learning when we instantiate them with different building blocks.

Further, compared to existing tuning methods where tuning structures are only attached to a subset of operations (\textit{e.g.,} only to MHA or only to FFN), our formulation can attach \tuner\ to all operations (MHA and FFN) or even to Transformer blocks, as can be seen in~\cref{fig:utuning}.

\subsection{Instantiations} \label{sec:instantiations}
The \utuning\ framework can be instantiated in two parts, \textit{i.e.,} the operations \uop\ existing in pre-trained Transformers with frozen parameters and the unified tuner \tuner\ with learnable parameters, respectively. To instantiate existing tuning methods, we can instantiate \utuning\ with corresponding \uop\ and \tuner. On top of that, we can generate various new tuning methods with different combinations of operations and tuners. 

\textbf{Operations. }
The emergence of parameter-efficient transfer learning is majorly motivated by the strong generalization capabilities of large foundation models pre-trained on a large corpus of data. Hence, on the micro level, we follow the practice of existing tuning methods and align the operations \uop\ term in our unified formulation with the design of pre-trained Transformers, which can be either MHA or FFN. This alignment allows the retainment of the generalization capabilities of the pre-trained Transformers. Different from the existing tuning paradigm for PETL, we additionally introduce a macro level to the operations, where the \uop\ term can also be a Transformer block. This further increases the flexibility of the formulation as well as the \utuning\ framework. The detailed instantiation of the operations can be viewed in~\cref{fig:utuning}.

\tuner. 
According to the instantiated operation, our \tuner\ can be connected in parallel to either MHA, FFN, or the block, as in~\cref{fig:utuning} (a). 
The instantiation of the \tuner\ is decoupled from the instantiation of the operations, which means the structure of the \tuner\ does not necessarily need to be consistent with the corresponding operation.
This is in contrast to the existing tuning methods, where the tuner performs a similar operation as the existing operations in Transformers, \textit{e.g.,} adapter MLP for FFN. 

For the instantiation of \tuner, we introduce three variants extended from adapter~\cite{houlsby2019adapter}, prefix tuning~\cite{li2021prefix}, and prompt tuning~\cite{lester2021pompt}, respectively. Specifically, we use the parallel form of the above methods that we derive in~\cref{sec:rethink}, \textit{i.e.,} parallel adapter, parallel prefix, and parallel prompt (as shown in~\cref{fig:utuning}), such that \tuner\ can be easily plugged into or removed from pre-trained Transformers with high flexibility, which facilitates the deployment of such large models in various downstream applications. Such design also allows for easy extension to new tuning methods when new operations or structures emerge. 

Another important difference between our \utuning\ framework and existing approaches is that the instantiation of our \tuner\ is not limited to only one \uop\ in one block. This means the distributions of all modules in pre-trained models can be adjusted to fit the new data distributions in various downstream tasks. We study the performance of our \utuning\ framework with one, two, and three \tuner\ instantiated in the experiments.

\begin{table*}[ht]
\caption{
Performance and parameter comparisons on FGVC. The bold font represents the best accuracy, and the underline represents the second best accuracy. ``Mean'' denotes the average accuracy of datasets. ``Params'' is the average tunable parameters needed for training.
}
\label{tab:fgvc}
\vskip 0.1in
\begin{center}
\begin{small}
\begin{sc}
\resizebox{\linewidth}{!}
{
\begin{tabular}{lccccccc}
\toprule
Method & CUB\_200\_2011 & NABirds & OxfordFlowers & StanfordCars & StanfordDogs & Mean & Params.(M) \\
\midrule
Fully fine-tuning &          87.3  &     82.7  &           98.8  &          {\bf 84.5}  &          89.4  &  88.54 &        85.98 \\
Linear probing   &          85.3  &     75.9  &           97.9  &          51.3  &          86.2  &  79.32 &         0.18 \\
\midrule
BitFit          &          88.4  &     84.2  &           98.8  &          79.4  &          \underline{91.2}  &  88.41 &         0.28 \\
Prefix          &          87.5  &     82.0  &           98.0  &          74.2  &          90.2  &  86.37 &         0.36 \\
Adapter          &          87.1  &     \underline{84.3}  &           98.5  &          68.6  &          89.8  &  85.67 &         0.41 \\
VPT-Shallow      &          86.7  &     78.8  &           98.4  &          68.7  &          90.7  &  84.62 &         0.25 \\
VPT-Deep         &          \underline{88.5}  &     84.2  &           \underline{99.0}    &          83.6  &          90.2  &  \underline{89.11} &         0.85 \\
\midrule
U-Tuning    &          {\bf89.16}  &     {\bf85.39}  &           {\bf99.15}    &          \underline{84.14}  &          {\bf92.07}  &  {\bf89.98}   &         0.36 \\
\bottomrule
\end{tabular}
}
\end{sc}
\end{small}
\end{center}
\vskip -0.2in
\end{table*}

\subsection{Analysis and Discussion}
\label{sec:analysis}
\textbf{Scaling.}
Our \utuning\ framework is able to attach the \tuner\ to all modules in pre-trained Transformers. To further improve the flexibility as well as to balance between different modules, we introduce a channel-wise scaling factor to all the parallel tuners, \textit{i.e.,}
\begin{equation}
\operatorname{\tuner}(\boldsymbol{x})^{\prime} = \boldsymbol{s} \cdot \operatorname{\tuner}(\boldsymbol{x}),
\end{equation}
where $\boldsymbol{s}$ is the channel-wise scaling factor. We also study other forms of scaling factors in experiments.

\textbf{Parameter analysis.}
The PETL methods fix most of the pre-trained model parameters and introduce only a few learnable parameters, as is our method. The total number of our tunable parameters basically depends on the current method used in \tuner\ . For example, the amount of parameters for VPT-Deep~\cite{jia2022vpt} is ${L}\times{n}\times{d}$, where $L$ is the total layers, $n$ is the number of prompts, and $d$ is the dimension of prompts. Our following experiments show that \utuning\ can surpass the method used by the full layer ($L$ layers) with a lower number of parameters, thus achieving a fewer parameter count by comparison. Other parameters have similar results, such as the number of prefix tokens, and the hidden layer dimension of the adapter.

\textbf{Diverse unified perspective.}
In NLP,~\citeauthor{he2022towards} also look at PETL from a unified view, aiming to establish close connections and unify to the adapter method. Although \utuning\ takes inspiration, our approach is different in motivations, design principles, and fields. \utuning\ derives existing methods in a completely equivalent form and proposes a unified paradigm in which any tuning method can be independent of the main branch, whereas~\citeauthor{he2022towards} unify PETL methods to find several variant adapters in an approximate manner.

\textbf{Expansibility and applicability.}
Our approach proposes a unified paradigm in which the two core operators, \uop\ and \tuner\, represent the submodules to be optimized and the tuning methods in parallel. Both operators need not be limited to the approach presented in this paper, but extend to other approaches, \textit{e.g.}, LoRA~\cite{hu2021lora}, and even newer ones. Furthermore, \utuning\ also allows for plug and play of \tuner\ that connects to the original modules through residual connection. In this way, rapid adaption and deployment are possible for all kinds of downstream tasks.

\begin{table}[t]
\caption{
Comparison between the \utuning\ framework and existing tuning approaches for PETL on CIFAR-100. \utuning\texttt{/1} and \utuning\texttt{/12} represent introducing \tuner\ in the first layer and in all layers of Vision Transformers respectively. $\dagger$ denotes results from our reimplementation. $\ast$ denotes tri-\tuner\ variant for \utuning.
}
\label{tab:CIFAR-100}
\vskip 0.1in
\begin{center}
\begin{small}
\begin{sc}
\tablestyle{8pt}{1.0}
\begin{tabular}{lcc}
\toprule
Method & CIFAR-100 & Params.(M) \\
\midrule
Fully fine-tuning &      89.12 &        86.04 \\
Linear probing     &      85.95 &         0.07 \\
\midrule
Prefix$^{\dagger}$      &      90.95 &         0.26 \\
VPT-Shallow$^{\dagger}$ &      86.62 &         0.08 \\
VPT-Deep$^{\dagger}$    &      91.58 &         0.17 \\
Adapter$^{\dagger}$     &      91.80 &         0.27 \\
AdaptFormer &      91.86 &         1.26 \\
\midrule
U-Tuning/1   &      91.86 &         0.11 \\
U-Tuning/12  &     \underline{92.57} &         0.59 \\
U-Tuning/12$^{\ast}$  &     \textbf{92.75} &         0.67 \\
\bottomrule
\end{tabular}
\end{sc}
\end{small}
\end{center}
\vskip -0.3in
\end{table}

\section{Experiments} \label{sec:exp}
In this section, we evaluate and analyze the performance and design of our proposed \utuning\ method on several downstream tasks. Specifically, we describe our experiment set up in~\cref{sec:setup}, compare with SOTA methods in~\cref{sec:sota}, and present comprehensive ablative analysis in~\cref{sec:study}.

\subsection{Experiment Setup} \label{sec:setup}
\textbf{Datasets.}
To evaluate the capability of our method on various downstream tasks, we take  CIFAR-100~\cite{krizhevsky2009cifar100} and FGVC datasets, where CIFAR-100 is typically used on general image classification tasks. FGVC is the benchmark of fine-grained visual classification tasks, which consists of CUB-200-2011~\cite{wah2011cub200}, NABirds~\cite{van2015nabirds}, Oxford Flowers~\cite{nilsback2008flowers}, Stanford Cars~\cite{gebru2017cars}, and Stanford Dogs~\cite{khosla2011dogs}. We report the Top-1 accuracy on CIFAR-100 and FGVC datasets for the experiments of transfer learning. 

\textbf{Baselines.}
We first compare our method with baselines in transfer learning and divide them into two categories: \textbf{(i)} Traditional methods: fully fine-tuning, which updates all the parameters of model, and linear probing, which freezes the pre-trained backbone and only tunes the classifier. \textbf{(ii)} PETL methods: BitFit~\cite{zaken2021bitfit}, adapter~\cite{houlsby2019adapter}, prefix~\cite{li2021prefix}, VPT~\cite{lester2021pompt}, and AdaptFormer~\cite{chen2022adaptformer}.

\textbf{Implementation details.}
If not specified, we experiment with ViT-B/16~\cite{dosovitskiy2020vit} model pre-trained on ImageNet-21K~\cite{deng2009imagenet} as conducted in VPT~\cite{lester2021pompt}. For most datasets, we preprocess the data with a random resized crop and a random horizontal flip to the size of $224\times224$. AdamW~\cite{loshchilov2017adamw} optimizer and cosine annealing learning rate scheduler are used with linear warm-up strategy. We use a combination of \tuner\ with a channel-wise scaling strategy instead of a single one to get better performance. More details are introduced in~\cref{app:exp}.

\subsection{Comparisons with the SOTA} \label{sec:sota}
We compare the transfer ability of our \utuning\ framework with different existing approaches for parameter-efficient transfer learning on CIFAR-100 and FGVC datasets, which are evaluated in Top-1 accuracy. 

\textbf{FGVC.}
As shown in~\cref{tab:fgvc}, our \utuning\ outperforms VPT~\cite{jia2022vpt} and other tuning methods (fully fine-tuning, linear probing, bias, prefix, and adapter) on the average accuracy of five FGVC datasets. For all FGVC datasets other than StanfordCars, our \utuning\ outperforms existing tuning methods by notable margins. For StanfordCars, the \utuning\ falls short of fully fine-tuning but performs stronger than other tuning methods. We suspect that it is because the intra-class difference is too small and hence more parameters are required for this dataset. The used construction form of \utuning\ for each dataset is listed in \cref{tab:fgvc_cons} in \cref{app:fgvc_cons}, which is the \utuning\ form with the best performance in all possible construction forms. The ablation studies on the performance for different construction forms of \utuning\ are discussed in \cref{sec:study}.

\textbf{CIFAR-100.}
As in~\cref{tab:CIFAR-100}, our approach goes beyond fully fine-tuning and linear probing by 3.45\% and 6.62\%, respectively. Compared with existing methods for PETL, \utuning\ achieves on-par or better performances. Specifically, \utuning\ with the first layer adapted by the unified tuner \tuner\ achieves similar performance to the previous best performances achieved by existing approaches (91.86 of both \utuning\texttt{/1} and AdaptFormer), while reducing the parameter requirement by over 10 times (0.11M of \utuning\texttt{/1} and 1.26M of AdaptFormer). When all layers are adapted by the unified tuner, we observe a notable improvement over state-of-the-art accuracy (92.57 of \utuning\texttt{/12} \textit{vs.} 91.86 of AdaptFormer). Note that, compared with AdaptFormer, \utuning\texttt{/12} only uses less than half of the parameters used by AdaptFormer (0.59M of \utuning\texttt{/12} \textit{vs.} 1.26M of AdaptFormer). Using the tri-\tuner\ variant, \utuning\texttt{/12} achieves the strongest result. 

\begin{table}[t]
\caption{
Verification of the equivalence between the original and the parallel forms of existing PETL methods.
}
\label{tab:equivalent}
\vskip 0.1in
\begin{center}
\begin{small}
\begin{sc}
\begin{tabular}{lccc}
\toprule
Type           & Adapter & Prefix & Prompt \\
\midrule
Original       & 91.80   & 90.95  & 91.19  \\
Parallel       & 91.86   & 91.01  & 91.26  \\
\bottomrule
\end{tabular}
\end{sc}
\end{small}
\end{center}
\vskip -0.2in
\end{table}

\begin{table}[t]
\caption{
Single \tuner\ with different parallel tuners attached to different operations \uop. 
}
\label{tab:single}
\vskip 0.1in
\begin{center}
\begin{small}
\begin{sc}
\setlength\tabcolsep{8pt}
\begin{tabular}{lccc}
\toprule
Method &   MHA   & FFN & Block  \\
\midrule
VPT    & 90.97 & - & - \\
\midrule
P-Adapter   &  \textbf{92.43} & 92.19 & 92.13 \\
P-Prefix    &  91.42 & 90.27  & 90.26 \\
P-Prompt    &  91.54 & 90.01 & 90.12 \\
\bottomrule
\end{tabular}
\end{sc}
\end{small}
\end{center}
\vskip -0.3in
\end{table}

\subsection{Ablation Study} \label{sec:study}
We present comprehensive ablative analysis here for a deeper understanding of \utuning.

\textbf{The equivalency of the original form and the parallel form of existing PETL methods.}
We first empirically verify the alignment in performance between the existing tuning methods in their original form and their parallel equivalent derivations. As in~\cref{tab:equivalent}, it can be observed that the performance gap is small enough to be neglected. This equivalence also indicates that our unified framework can effectively encompass existing tuning methods for PETL.

\textbf{Different combinations of \uop\ and \tuner.}
After showing that our \utuning\ framework can encompass existing tuning methods for PETL, we further explore various combinations of \uop\ and \tuner\ to derive new tuning methods. We focus our explorations here on CIFAR-100 dataset. 

\textbf{(i)} Single \tuner. 
We start by using one \tuner\ per block, which is similar to existing PETL methods. We explore different \tuner\ forms on different operations \uop. Given the existing paradigms for PETL, we focus our exploration on the aforementioned parallel adapter (P-Adater), parallel prefix (P-Prefix), and parallel prompt (P-Prompt). As in \cref{tab:single}, we connect them to MHA, FFN, and the whole block, respectively. It can be observed that connecting P-Adapter to MHA achieves the strongest result, which means that the existing practice is not necessarily the optimal combination. Further, compared to VPT~\cite{jia2022vpt}, our single \tuner\ variant can already achieve a notable improvement. 

\begin{table}[t]
\caption{
Dual-\tuner\ with different tuner instantiations attached to MHA and FFN.
}
\label{tab:two}
\vskip 0.1in
\begin{center}
\begin{small}
\begin{sc}
\setlength\tabcolsep{6pt}
\begin{tabular}{lcccc}
\toprule
\diagbox{MHA}{FFN}   &   P-Adapter &   P-Prefix &   P-Prompt \\
\midrule
P-Adapter    &     {\bf 92.57} &    92.45 &    92.30 \\
P-Prefix     &     92.10 &    91.52 &    91.69 \\
P-Prompt     &     92.27 &    91.60 &    91.76  \\
\bottomrule
\end{tabular}
\end{sc}
\end{small}
\end{center}
\vskip -0.1in
\end{table}

\begin{table}[t]
\caption{
Tri-\tuner\ with different tuner instantiations for the block. The parallel adapter is used for MHA and FFN.
}
\label{tab:triple}
\vskip 0.1in
\begin{center}
\begin{small}
\begin{sc}
\begin{tabular}{lccc}
\toprule
Block          & P-Adapter & P-Prefix & P-Prompt \\
\midrule
Accuracy       & 92.69   & \textbf{92.75} &  92.53 \\
\bottomrule
\end{tabular}
\end{sc}
\end{small}
\end{center}
\vskip -0.1in
\end{table}

\textbf{(ii)} Dual-\tuner. 
We further explore combinations of \uop\ and \tuner\ when both MHA and FFN are adapted by the \tuner\ in \cref{tab:two}. Interestingly, the \utuning\ framework performs the best when we use parallel adapters for both MHA and FFN. Further, we found the performance with at least one \tuner\ using the parallel adapter is generally higher. Dual-\tuner\ also outperforms the single \tuner\ variant, indicating that all modules in pre-trained Transformers should be adapted for PETL. Unless otherwise specified, we use dual-\tuner\ as our default version for the remaining of the ablative analysis.

\textbf{(iii)} Tri-\tuner. 
On top of the best dual-\tuner, we add an additional block \tuner\ in \cref{tab:triple}. The performances of adding the parallel prefix and the parallel adapter are similar. Compared with dual-\tuner, tri-\tuner\ further improves by 0.18\%.

\begin{figure}[t]
\vskip 0.1in
\begin{center}
\centerline{\includegraphics[width=0.95\columnwidth]{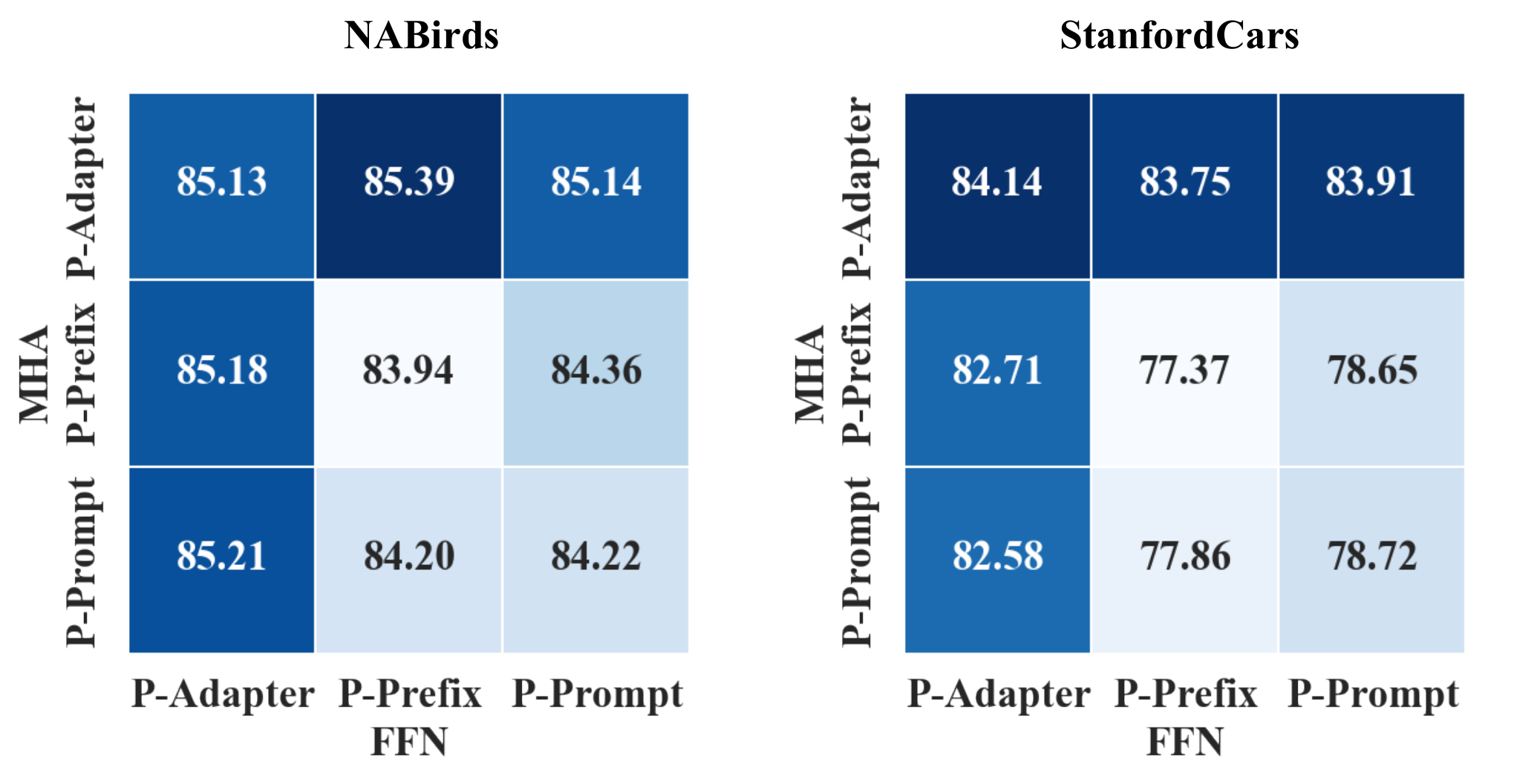}}
\caption{
Generalization of our findings in dual-\tuner\ to NABirds and StanfordCars.
}
\label{fig:generalization}
\end{center}
\vskip -0.2in
\end{figure}

\begin{table}[t]
\caption{
Ablation studies on the scaling strategies. 
}
\label{tab:joint}
\vskip 0.1in
\begin{center}
\begin{small}
\begin{sc}
\setlength\tabcolsep{20pt}
\begin{tabular}{lc}
\toprule
Type        &   Accuracy \\
\midrule
Direct connection    & 92.18 \\
Scalar scaling       & 92.22 \\
Channel-wise scaling & {\textbf{92.57}} \\
Input-dependent scaling  & 92.12 \\
\bottomrule
\end{tabular}
\end{sc}
\end{small}
\end{center}
\vskip -0.2in
\end{table}

\begin{figure}[t]
\vskip 0.1in
\begin{center}
\centerline{\includegraphics[width=\columnwidth]{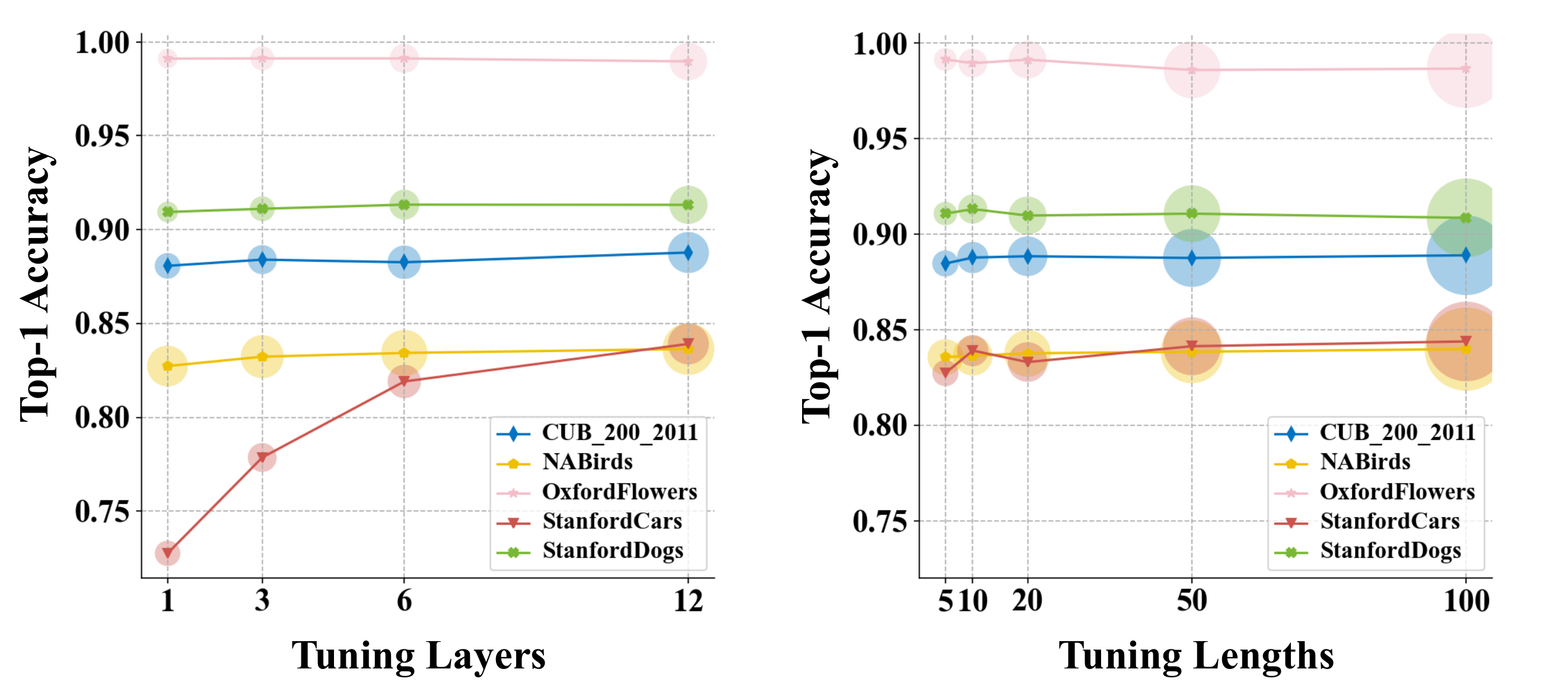}}
\caption{
Ablation studies on the tuning layers and the length of dimension for \utuning\ on FGVC.
}
\label{fig:layers_lengths}
\end{center}
\vskip -0.3in
\end{figure}

\begin{table}[t]
\caption{
Ablative analysis on different pre-trained Transformers.
}
\label{tab:backbone}
\vskip 0.1in
\begin{center}
\begin{small}
\begin{sc}
\begin{tabular}{lcccc}
\toprule
\multicolumn{1}{l|}{Architecture}   & \multicolumn{3}{c|}{ViT/B-16} & \multicolumn{1}{l}{ViT/L-14} \\
\multicolumn{1}{l|}{Pre-training}    & IN-21K & MAE   & \multicolumn{1}{c|}{CLIP}  & CLIP                         \\
\midrule
\multicolumn{1}{l|}{Adapter}        & 91.88        & 86.11 & \multicolumn{1}{c|}{89.00} & 92.01                        \\
\multicolumn{1}{l|}{Prefix}         & 90.95        & 78.35 & \multicolumn{1}{c|}{86.54} & 90.50                         \\
\multicolumn{1}{l|}{VPT-Shallow} & 86.62        & 55.98 & \multicolumn{1}{c|}{85.47} & 89.51                        \\
\multicolumn{1}{l|}{VPT-Deep}    & 91.58        & 82.95 & \multicolumn{1}{c|}{87.77} & 91.56                        \\
\multicolumn{1}{l|}{U-Tuning}        & \textbf{92.57}       & \textbf{86.17} & \multicolumn{1}{c|}{\textbf{89.17}}  & \textbf{92.38}     \\
\bottomrule
\end{tabular}
\end{sc}
\end{small}
\end{center}
\vskip -0.3in
\end{table}

\textbf{Generalization of our findings to fine-grained datasets.} We further experiment with dual-\tuner\ on fine-grained datasets to examine our previous findings on CIFAR-100. The results are shown in~\cref{fig:generalization}. On fine-grained datasets such as NABirds and StanfordCars, the observations are consistent: models with adapters as the \tuner\ generally achieve stronger performance.

\textbf{Different scaling strategies.}
In~\cref{tab:joint}, we compare the default channel-wise scaling with other scaling strategies, \textit{i.e.,} direct connection (no scaling), scalar scaling, and input-dependent scaling (SE~\cite{hu2018senet}-like). We find channel-wise scaling better than other variants.

\textbf{Different number of layers with \tuner.}
In~\cref{fig:layers_lengths}, we vary the number of layers to insert the \tuner\ from bottom to top on the FGVC dataset. Increasing the number of layers with \tuner\ effectively raises the performance of StanfordCars, while the other datasets are less affected.

\textbf{Varying the length of dimensions.}
The length of dimensions represents the hidden dimension for parallel adapters and the number of prompt/prefix tokens. Altering the length of dimensions affects performances and the number of parameters simultaneously. In~\cref{fig:layers_lengths}, we show that increasing the length of dimensions has a relatively great impact when its value is small. 

\textbf{Different pre-training sources and backbones.}
In~\cref{tab:backbone}, we compare several tuning methods with our proposed \utuning\ based on different architectures and pre-trained models, \textit{i.e.,} IN-21K~\cite{deng2009imagenet}, MAE~\cite{he2022mae} and CLIP~\cite{radford2021clip} pre-trained ViT/B-16~\cite{dosovitskiy2020vit}, as well as CLIP pre-trained ViT/L-14. The \utuning\ framework achieves the best performance for all pre-trained Transformers.

\section{Conclusion} \label{sec:conclusion}
In this work, we revisit mainstream methods for parameter-efficient transfer learning (PETL) and establish a unified framework for PETL. Our approach, dubbed \utuning, is composed of an operation \uop\ with frozen parameters and a unified tuner \tuner\ for adapting the operation to various downstream tasks. Instantiated with various building blocks, our \utuning\ framework can encompass most existing tuning methods for PETL and meanwhile generate new tuning methods, such that \utuning\ is more flexible and efficient in the training and deployment process. Empirically, we show that the \utuning\ paradigm achieves strong results on transfer learning. We hope that \utuning\ can provide a new perspective on parameter-efficient tuning methods.

\bibliography{UTuning}
\bibliographystyle{icml2023}

\newpage
\appendix
\onecolumn
\section{Appendix}

\subsection{Detailed derivations}
\label{app:form}
\begin{itemize}
\item \textbf{Prefix tuning}: The following is the detailed derivation of \cref{eq:prefix_parallel}.

\begin{equation}
\label{eq:prefix_detaied_derivation}
\begin{array}{l}
\text { head }=\operatorname{Attn}\left(\boldsymbol{x} \boldsymbol{W}_{q}, \operatorname{concat}\left(\boldsymbol{K}_{pre}, \boldsymbol{x} \boldsymbol{W}_{k}\right), \operatorname{concat}\left(\boldsymbol{V}_{pre}, \boldsymbol{x} \boldsymbol{W}_{v}\right)\right) \\
=\operatorname{softmax}\left(\boldsymbol{x} \boldsymbol{W}_{q} \operatorname{concat}\left(\boldsymbol{K}_{pre}, \boldsymbol{x} \boldsymbol{W}_{k}\right)^{\top}\right)\left[\begin{array}{c}
\boldsymbol{V}_{pre} \\ 
\boldsymbol{x} \boldsymbol{W}_{v}
\end{array}\right] \\ 
=(1-\lambda(\boldsymbol{x})) \operatorname{softmax}\left(\boldsymbol{x} \boldsymbol{W}_{q} \boldsymbol{W}_{k}^{\top} \boldsymbol{x}^{\top}\right) \boldsymbol{x} \boldsymbol{W}_{v}+\lambda(\boldsymbol{x}) \operatorname{softmax}\left(x \boldsymbol{W}_{q} \boldsymbol{K}_{pre}^{\top}\right) \boldsymbol{V}_{pre} \\ \\
=(1-\lambda(\boldsymbol{x})) \operatorname{Attn}\left(\boldsymbol{x} \boldsymbol{W}_{q}, \boldsymbol{x} \boldsymbol{W}_{k}, \boldsymbol{x} \boldsymbol{W}_{v}\right)+\lambda(\boldsymbol{x}) \operatorname{Attn}\left(\boldsymbol{x} \boldsymbol{W}_{q}, \boldsymbol{K}_{pre}, \boldsymbol{V}_{pre}\right) \\ \\
=(1-\lambda({\boldsymbol{Q}, \boldsymbol{K}, \boldsymbol{K}_{pre}})) \underbrace{\operatorname{Attn}\left(\boldsymbol{Q}, \boldsymbol{K}, \boldsymbol{V}\right)}_{\text {standard attention }}+\lambda(\boldsymbol{Q}, \boldsymbol{K}, \boldsymbol{K}_{pre}) \underbrace{\operatorname{Attn}\left(\boldsymbol{Q}, \boldsymbol{K}_{pre}, \boldsymbol{V}_{pre}\right)}_{\text {independent of } \boldsymbol{K}_{pre}, \boldsymbol{V}_{pre}}
\end{array}
\end{equation}

where 
\begin{equation}
\lambda(\boldsymbol{Q}, \boldsymbol{K}, \boldsymbol{K}_{pre}) =\frac{\sum_{i} \exp \left( \boldsymbol{Q} \boldsymbol{K}_{pre}^{\top} \right)_{i}}{\sum_{i} \exp \left(\boldsymbol{QK^{\top}}\right)_{i}+\sum_{j} \exp \left(\boldsymbol{Q} \boldsymbol{K}_{pre}^{\top}\right)_{j}}, 
\end{equation}

\item \textbf{Prompt tuning}: The following is the detailed derivation of~\cref{eq:prompt_parallel}.

\begin{equation}
\label{eq:prompt_detaied_derivation}
\begin{array}{l}
\text { head }=\operatorname{Attn}\left(\operatorname{concat}\left(\boldsymbol{x}, \boldsymbol{x}_{pro} \right) \boldsymbol{W}_{q}, \operatorname{concat}\left(\boldsymbol{x}, \boldsymbol{x}_{pro} \right) \boldsymbol{W}_{k}, \operatorname{concat}\left(\boldsymbol{x}, \boldsymbol{x}_{pro} \right) \boldsymbol{W}_{v} \right) \vspace{3ex}\\
=\operatorname{concat}\left( \operatorname{softmax}\left(\boldsymbol{x} \boldsymbol{W}_{q} \operatorname{concat}\left(\boldsymbol{x} \boldsymbol{W}_{k}, \boldsymbol {x}_{pro} \boldsymbol{W}_{k} \right)^{\top}\right)\left[\begin{array}{c} \boldsymbol{x} \boldsymbol{W}_{v} \\ \boldsymbol{x}_{pro} \boldsymbol{W}_{v} \end{array}\right], \right. \vspace{1.5ex}\\
\left. \hspace{1.6cm}  \operatorname{softmax}\left(\boldsymbol{x}_{pro} \boldsymbol{W}_{q} \operatorname{concat}\left(\boldsymbol{x} \boldsymbol{W}_{k}, \boldsymbol{x}_{pro} \boldsymbol{W}_{k} \right)^{\top}\right)\left[\begin{array}{c} \boldsymbol{x} \boldsymbol{W}_{v} \\ \boldsymbol{x}_{pro} \boldsymbol{W}_{v} \end{array}\right] \right)  \vspace{3ex}\\
=\operatorname{concat}\left(( 1-\lambda({\boldsymbol{Q}, \boldsymbol{K}, \boldsymbol{K}_{pro}})) \operatorname{Attn}\left(\boldsymbol{Q}, \boldsymbol{K}, \boldsymbol{V}\right)+\lambda(\boldsymbol{Q}, \boldsymbol{K}, \boldsymbol{K}_{pro}) \operatorname{Attn}\left(\boldsymbol{Q}, \boldsymbol{K}_{pro}, \boldsymbol{V}_{pro}\right),
\right. \vspace{1.5ex}\\
\left. \hspace{1.6cm} (1-\beta({\boldsymbol{Q}_{pro}, \boldsymbol{K}_{pro}, \boldsymbol{K}})) \operatorname{Attn}\left(\boldsymbol{Q}_{pro}, \boldsymbol{K}_{pro}, \boldsymbol{V}_{pro}\right)+\beta(\boldsymbol{Q}_{pro}, \boldsymbol{K}_{pro}, \boldsymbol{K}) \operatorname{Attn}\left(\boldsymbol{Q}_{pro}, \boldsymbol{K}, \boldsymbol{V} \right)
\right)
\end{array}
\end{equation}

where 
\begin{equation}
\lambda(\boldsymbol{Q}, \boldsymbol{K}, \boldsymbol{K}_{pro}) =\frac{\sum_{i} \exp \left( \boldsymbol{Q} \boldsymbol{K}_{pro}^{\top} \right)_{i}}{\sum_{i} \exp \left(\boldsymbol{QK^{\top}}\right)_{i}+\sum_{j} \exp \left(\boldsymbol{Q} \boldsymbol{K}_{pro}^{\top}\right)_{j}}, 
\end{equation}

\begin{equation}
\beta(\boldsymbol{Q}_{pro}, \boldsymbol{K}_{pro}, \boldsymbol{K}) =\frac{\sum_{i} \exp \left( \boldsymbol{Q}_{pro} \boldsymbol{K}^{\top} \right)_{i}}{\sum_{i} \exp \left(\boldsymbol{Q}_{pro} \boldsymbol{K}_{pro}^{\top}\right)_{i}+\sum_{j} \exp \left(\boldsymbol{Q}_{pro} \boldsymbol{K}^{\top}\right)_{j}}, 
\end{equation}

\end{itemize}

\subsection{Experimental Details} \label{app:exp}
\subsubsection{Datasets}
\label{app:data}
We list the description of each dataset and our experimental settings in~\cref{tab:dataset}, which includes the number of classes and the amount of images in training set and test set.

\begin{table}[h]
\caption{Datasets and settings in our experiments.}
\label{tab:dataset}
\vskip 0.15in
\begin{center}
\begin{small}
\begin{sc}
\begin{tabular}{llll}
\toprule
Dataset & Classes & Train & Test  \\
\midrule
General Image Classification \\
CIFAR-100~\cite{krizhevsky2009cifar100} & 100 & 50000 & 10000  \\
\midrule
Fine-grained Visual Classification (FGVC) \\ 
CUB\_200\_2011~\cite{wah2011cub200} & 200 & 5994 & 5794  \\
NABirds~\cite{van2015nabirds} & 555 & 23929 & 24633  \\
OxfordFlowers~\cite{nilsback2008flowers} & 102 & 1020 & 6149  \\
StanfordCars~\cite{gebru2017cars} & 196 & 8144 & 8041  \\
StanfordDogs~\cite{khosla2011dogs} & 120 & 12000 & 8580  \\
\bottomrule
\end{tabular}
\end{sc}
\end{small}
\end{center}
\vskip -0.1in
\end{table}
\subsubsection{Hyperparameters}
\label{app:hyper}

To help reproduce the experiments conducted in this paper, we list all the hyperparameters and the architecture's units. Concretely, the training hyperparameters are listed in the first section of \cref{tab:hyperparameters}. In addition, the choices of the other attributes, such as the number of Transformer layers, the length of dimensions, the used OP of transformer, the basic tuners and so on, are listed in the others sections of \cref{tab:hyperparameters}, respectively.

\begin{table}[h]
\caption{Hyperparameters and architecture's units in detail}
\label{tab:hyperparameters}
\vskip 0.15in
\begin{center}
\begin{small}
\begin{sc}
\begin{tabular}{llll}
\toprule
Config & Value \\
\midrule
Batch size   & 32  \\ 
Optimizer            & AdamW~\cite{loshchilov2017adamw}  \\
Weight decay         & 0.05   \\ 
Base learning rate range  & \{0.001, 0.005\}  \\
Learning rate schedule  & cosine decay  \\
Training epochs range   & \{50, 100\}  \\
Warmup epochs &   10 \\
Augmentation  &   RandomResizedCrop, RandomHorizontalFlip\\ 
\midrule
The number of Transformer layers  & \{1, 1-3, 1-6, 1-12\}  \\ 
The length of dimensions & \{5, 10, 20, 50, 100\}  \\ 
\midrule
OP & \makecell[l]{MHA~\cite{vaswani2017attention} \\ FFN~\cite{vaswani2017attention} \\ Block~\cite{vaswani2017attention}} \\
\midrule
U-Tuner & \makecell[l]{Adapter~\cite{houlsby2019adapter} \\   Prefix~\cite{li2021prefix} \\ Prompt~\cite{jia2022vpt}} \\
\midrule
Scaling strategies & \makecell[l]{direct connection,\\ scalar scaling, \\ channel-wise scaling, \\ input-dependent scaling} \\
\midrule
Architecture & \makecell[l]{ViT/B-16~\cite{dosovitskiy2020vit} \\ ViT/B-14~\cite{dosovitskiy2020vit}}  \\ 
\midrule
Pre-trained & \makecell[l]{ImageNet-21K~\cite{deng2009imagenet} \\ MAE~\cite{he2022mae} \\ CLIP~\cite{radford2021clip}}  \\
\bottomrule
\end{tabular}
\end{sc}
\end{small}
\end{center}
\vskip -0.1in
\end{table}

\subsubsection{The construction forms of \utuning\ on FGVC Datasets} \label{app:fgvc_cons}

As described in \cref{app:hyper}, we conduct experiments on different construction forms of \utuning\ by attaching the \tuner\ units to the OP units, MHA and FFN, respectively. \cref{tab:fgvc_cons} lists the construction forms for each FGVC dataset, which achieves the best performance in all the experiments above. We find that there is a discrepancy in the best construction form of \utuning\ for different datasets. For example, CUB\_200\_2011 benefits from the \utuning\ constructed with prompt on MHA and adapter on FFN, while OxfordFlowers prefers the \utuning\ constructed with adapter on MHA and prompt on FFN. This phenomenon further demonstrates that our proposed \utuning\ paradigm offers the shortcuts for the downstream tasks adaption of the pre-trained model.

\begin{table*}[t]
\caption{The construction forms of \utuning\ on FGVC datasets.}
\label{tab:fgvc_cons}
\vskip 0.15in
\begin{center}
\begin{small}
\begin{sc}
\resizebox{\linewidth}{!}
{
\begin{tabular}{lccccccc}
\toprule
{\bf OP} & CUB\_200\_2011 & NABirds & OxfordFlowers & StanfordCars & StanfordDogs \\
\midrule
{\bf MHA} &  Prompt  &   Adapter  &   Adapter   &    Adapter  &          Prefix \\
{\bf FFN} &  Adapter  &   Prefix  &   Prompt    &    Adapter  &          Adapter \\
\bottomrule
\end{tabular}
}
\end{sc}
\end{small}
\end{center}
\vskip -0.1in
\end{table*}

\end{document}